\documentclass[final]{cvpr}

\usepackage{times}
\usepackage{epsfig}
\usepackage{graphicx}
\usepackage{amsmath}
\usepackage{amssymb}

\usepackage{multicol}
\usepackage{float}
\usepackage{mathtools}
\usepackage{amsmath}
\newcommand{\norm}[1]{\lVert#1\rVert_2}
\usepackage{algorithm}
\usepackage[noend]{algpseudocode}
\usepackage[font=small,labelfont=bf]{caption}
\usepackage{multirow}
\usepackage{graphicx}
\usepackage{grffile}
\usepackage{booktabs}
\usepackage{adjustbox}
\usepackage{diagbox}
\newcommand{\revisionPQ}[2]{{\color[rgb]{0,0,0}#2}}
\newcommand{\revisionRG}[2]{{\color[rgb]{0,0,0}#2}}
\usepackage{subcaption}
\usepackage{placeins}
\usepackage[symbol]{footmisc}

\usepackage[pagebackref=true,breaklinks=true,colorlinks,bookmarks=false]{hyperref}

\def\cvprPaperID{577} 
\def\confYear{CVPR 2021}

\begin{document}

\title{Towards Imperceptible Query-limited Adversarial Attacks with Perceptual Feature Fidelity Loss}

\author{Pengrui Quan\thanks{denotes equal contribution}\\
University of California, Los Angeles\\
Los Angeles, U.S.\\
{\tt\small prquan@g.ucla.edu}
\and
Ruiming Guo$^*$\\
The Chinese University of Hong Kong\\
Shatin, Hong Kong\\
{\tt\small greenming@link.cuhk.edu.hk}
\and 
Mani Srivastava\\
University of California, Los Angeles\\
Los Angeles, U.S.\\
{\tt\small mbs@ucla.edu}
}

\maketitle



	\begin{figure*}[h]
        \centering \includegraphics[width=0.85\linewidth]{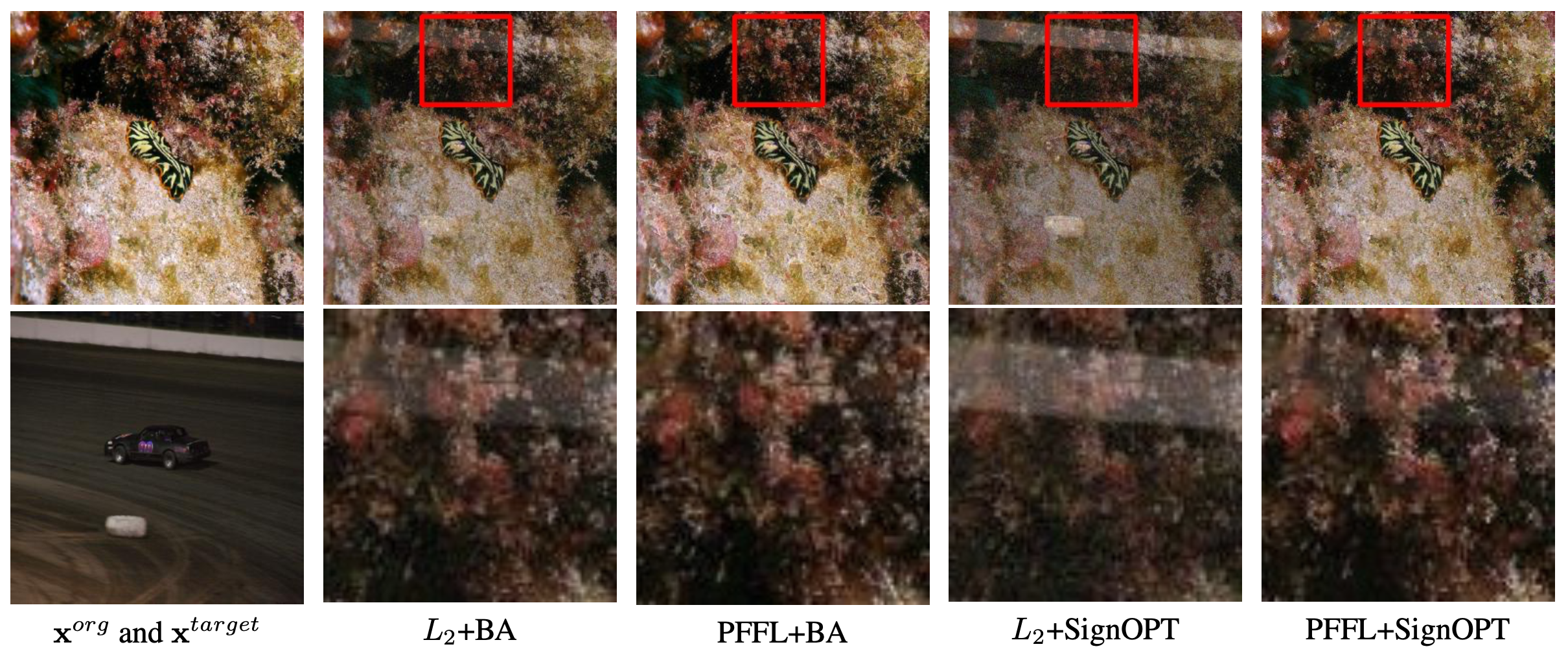}\par
		\caption{Visual comparison between different objectives. From left to right, first row: original image $\mathbf{x}^{org}$ and adversarial images generated using $L_2$ + Boundary Attack \cite{brendel2017decision}, PFFL (ours) + Boundary Attack, $L_2$ + SignOPT \cite{cheng2019sign}, and PFFL (ours) + SignOPT; Second row: image in the targeted class $\mathbf{x}^{target}$ and zoomed patches of each adversarial image. Our methods (PFFL+BA and PFFL+SignOPT) can visibly suppress the ghosting effects in the generated adversarial image with the same number of queries to blackbox model.}\label{direct_compare}
		\vspace{-.5em}
	\end{figure*}

\begin{abstract}
  Recently, there has been a large amount of work towards fooling deep-learning-based classifiers, particularly for images, via adversarial inputs that are visually similar to the benign examples. However, researchers usually use $L_p$-norm minimization as a proxy for imperceptibility, which oversimplifies the diversity and richness of real-world images and human visual perception. In this work, we propose a novel perceptual metric utilizing the well-established connection between the low-level image feature fidelity and human visual sensitivity, where we call it \emph{Perceptual Feature Fidelity Loss}. We show that our metric can robustly reflect and describe the imperceptibility of the generated adversarial images validated in various conditions. Moreover, we demonstrate that this metric is highly flexible, which can be conveniently integrated into different existing optimization frameworks to guide the noise distribution for better imperceptibility. The metric is particularly useful in the challenging black-box attack with limited queries, where the imperceptibility is hard to achieve due to the non-trivial perturbation power.
\end{abstract}

\section{Introduction}
	\subsection{Vulnerability of deep learning model}
    \revisionRG{}{In many areas, }it has been widely observed that deep neural networks are susceptible to adversarial \revisionRG{inputs}{image perturbations} \cite{szegedy2013intriguing,goodfellow2014explaining, alzantot2019genattack}. For instance, with \revisionRG{arg1}{visually} imperceptible perturbation added to the input images, deep-learning-based image classifiers \revisionRG{can}{could} be compromised \cite{carlini2017towards,ilyas2018black,cheng2018query,brendel2017decision}. Generating adversarial inputs with inconspicuous perturbation is of vital importance to study the model robustness, adversarial defense, and  machine perception \revisionRG{arg1}{mechanism}. A commonly \revisionRG{agreed}{agreed-upon} assumption is that a small $L_p$-distance is a good proxy of the imperceptibility in the image domain, and thus many works seek to generate adversarial \revisionRG{inputs}{examples} with bounded $L_p$ distance \cite{croce2019sparse, cheng2018query, cheng2019sign, chen2020hopskipjumpattack}. However, recent works \cite{sen2019should, sharif2018suitability, wong2019wasserstein} reveal that visually imperceptible perturbations do not necessarily require small $L_p$-distance, which motivates us to deeply explore the intrinsic connection of perturbation between statistical distribution and imperceptibility. 
    
    First and foremost, our work starts from questioning the validity of treating every image pixel with equal importance in the $L_p$-based methods \cite{cheng2019sign, chen2020hopskipjumpattack}. A widely \revisionRG{acknowledge}{recognized and accepted conclusion} in Image Quality Assessment (IQA) is that human visual perception, as a highly complex and subjective system, does not have a uniform and consistent sensitivity in every pixel, which is very likely to be influenced by the local image statistics, such as intensity contrast, local structure consistency, and self-similarity. \revisionRG{arg1}{In the field of image processing, very often people associate human visual sensitivity with low-level image feature fidelity \cite{rad2019srobb}, \cite{lin2011perceptual}, e.g., most people are more sensitive to the distortion on smooth regions rather than the fast-changing textures. }Using \revisionRG{}{this empirical fact, we manage to guide the spatial distribution of the adversarial perturbations by minimizing the Perceptual Feature Fidelity Loss (PFFL), to make it consistent with the benign image statistics. This effectively allows us to ``entangle'' large adversarial perturbation in image spatial space with fewer query resources, through a manner hidden from the human visual attention, as shown in fig.~\ref{direct_compare}.} This is of particular use in query-limited adversarial attacks, where the noise level is relatively large due to insufficient iterations, thence making the noise consistent with low-level image statistics is a sensible way to achieve imperceptibility.
    \subsection{Imperceptibility of perturbation on images}
	\begin{figure}[h]
    \minipage{0.15\textwidth}
        SSIM=1
        \includegraphics[width=\linewidth]{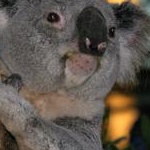} \centering original image\\\par
    \endminipage\hfill
    \minipage{0.15\textwidth}
        SSIM=0.43
        \includegraphics[width=\linewidth]{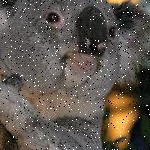} \centering PSNR=19.46\par
    \endminipage\hfill
    \minipage{0.15\textwidth}
        SSIM=0.94
        \includegraphics[width=\linewidth]{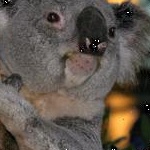} \centering PSNR=19.46\par
    \endminipage\hfill
		\caption{The visibility of different noise distribution. The right image apparently has better visual quality than the middle one. This illustrates that \romannumeral 1) large distortion remains imperceptible as it is consistent with low-level image feature; \romannumeral 2) simply referring to one metric could be unreliable.}
	\label{fig:ssim vs psnr}
	\end{figure}
    Perturbing the initial image ``unconsciously'' is quite valuable in many deep-learning related applications. The adversarial image attack is expected to be blocked and prevented by human inspectors or statistical tests \cite{metzen2017detecting, grosse2017statistical} if the perturbation is visually noticeable. Fig.~\ref{fig:ssim vs psnr} motivates the use of Perceptual Feature Fidelity Loss for guiding the spatial distribution of the adversarial image perturbations. To be more specific, the same amount of noise (Peak Signal-to-Noise Ratio: $19.46\mbox{dB}$) has been put on the same original image (left), which interestingly gives rise to two distinct perturbed images: the noise in the middle example is randomly and uniformly distributed in image space; while the noise in the right example is carefully designed with close correlation to the initial image statistics, i.e. low-level feature fidelity. Obviously, the resultant perturbations are only noticeable in the randomly-distributed example (middle) with significant grain effects, despite the fact that noise levels of both images remain identical. The reason behind this is that image feature fidelity perceived by the human eye does not change uniformly over distance in image space. In other words, relatively small perturbations in image space may result in large distortion in image features. Conversely, relatively large image space changes may still be imperceptible as long as the added perturbation causes small distortion of the initial image features.
    
    On the other hand, fig.~\ref{fig:ssim vs psnr} also reveals that it may not be reliable to evaluate the generated image quality only depending on a single indicator. In image and video processing tasks, people often compare the PSNR and SSIM metrics simultaneously to make a comprehensive evaluation of image quality \cite{dong2012nonlocally,buades2005non,wang2019cfsnet}. Given the difficulty of precisely balancing these two metrics in practical applications, this inspires us to propose a new metric to cover both PSNR and SSIM as much as possible to guide the imperceptible adversarial attacks effectively.
	
\subsection{Contribution}
    Although several meaningful attempts try to achieve imperceptible attack via exploiting image structural information \cite{luo2018towards, croce2019sparse, zhang2019smooth}, they essentially do not investigate the key ingredients of image structures that directly influence the final visual quality. As a result, this gives rise to a relatively rough manner of structure information utilization. Besides, producing imperceptible adversarial perturbation in black-box and hard-label setting (only access the top-1 category) is usually achieved via intensive queries, while improving noise distribution to reduce visibility in this challenging problem has not been explored before. In this paper, we propose a new metric, Perceptual Feature Fidelity Loss (PFFL) to improve the visual imperceptibility of generated adversarial example. The metric is based on the distortion of low-level image features and is flexibly applicable to various image contexts. We summarize our contribution as follows:

\begin{itemize}
    \item An in-depth study on creating adaptive and flexible adversarial attacks that achieve better perceptual quality with limited query budget (Fig. \ref{direct_compare}).
	\item Our experiments demonstrate that the proposed metric can effectively guide the noise spatial distribution so as to make it consistent with the initial image statistics, which has not been explored in the query-limited hard-label black-box attacks before.
	\item Novel feature distortion metric: the proposed metric takes both the overall noise power and the local feature distortion into consideration, quantitatively ensuring the comprehensive and reliable image quality evaluation.
\end{itemize}

\section{Background}\label{background}
    \noindent\textbf{Image quality assessment (IQA).} IQA \revisionRG{capture}{aims at characterizing} visual difference between two images captured by humans as closely as possible. Over decades, people have actively studied this problem with many meaningful attempts, such as PSNR, SSIM, BRISQUE \cite{huynh2008scope,hore2010image,mantiuk2011hdr,neiva2003desmame,watson1998toward}. However, designing a comprehensive and reliable evaluation indicator applicable to most image/video processing tasks remains intractable \cite{wang2004image, xu2017no}. As a compromise, in practice, researchers and practitioners usually compare two most-often used indicators simultaneously, i.e., PSNR and SSIM, to obtain a case-specific evaluation result empirically. The reason is that PSNR describes the overall noise power yet ignoring the local noise behavior; conversely, the SSIM indicator is of complex expression characterizing the local consistency of noise distribution yet neglecting the overall noise power. Hence, it is expected that combing them together can give a more comprehensive and accurate evaluation. 
    
\begin{figure}[h]
    \minipage{0.15\textwidth}
        \includegraphics[width=\linewidth]{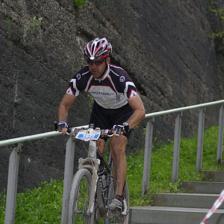} \centering original image\par
    \endminipage\hfill
    \minipage{0.15\textwidth}
        \includegraphics[width=\linewidth]{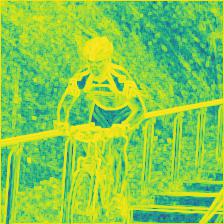} \centering local variance\cite{luo2018towards}\par
    \endminipage\hfill
    \minipage{0.15\textwidth}
        \includegraphics[width=\linewidth]{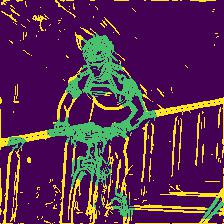} \centering our method\par
    \endminipage\hfill
    \vspace{-0.5em}
		\caption{Image directionality responses. Local variance is too sensitive to differentiate low-level image features (dark region corresponds to smooth area and bright region denotes edge or texture).}
	\label{fig:luo}
	\end{figure}

	\revisionPQ{}{\noindent\textbf{Perception-based adversarial attack.} With the growing awareness that the commonly-used $L_p$ distance oversimplifies human perception \cite{sen2019should, sharif2018suitability}, researchers have attempted to address the imperceptibility of adversarial attacks using different approaches. One direction is preserving semantical similarity \cite{hosseini2018semantic, engstrom2017rotation, zhao2017generating, qiu2019semanticadv}. However, since assessing the image fidelity and semantic similarity is still an unsolved problem \cite{wang2004image, xu2017no}, the above direction is not the focus of this work. Besides, other methods are trying to make improvement upon $L_p$-distance based on the conventional IQA metrics, such as a combination of $L_0$ and $L_\infty$ \cite{croce2019sparse}, local smoothness \cite{zhang2019smooth}, space saturation \cite{zhao2019towards} and local variance \cite{luo2018towards}.} However, these methods still lack in-depth investigation of the predominant factors affecting image visual quality. This gives rise to the rough image feature maps. For instance, as shown in fig.~\ref{fig:luo}: compared to our refined feature response (right), the common response map used in \cite{luo2018towards} is much more susceptible to image outlier, blurring, and background interference. Besides, nearly all of the above works target at white-box adversarial attacks, where imperceptible perturbation can have already been produced given the access to victim models. \revisionPQ{}{In this work, we study a more challenging attack setting, the hard-label black-box attack, where the query budget could become the restriction from reducing noise power. Given the relatively large perturbation, leveraging the low-level image feature consistency is desirable to achieve imperceptibility.}
	
    \noindent\textbf{Perception on low-level image feature.} The human visual \revisionRG{evaluation}{perception} is a quite sophisticated system related to many aspects, e.g., image resolution, object types, image contrast, user preference, etc. \cite{wu2019survey, haines1992effects, hudson2018jpeg}. \revisionRG{arg1}{An empirical consensus in practice is that the image visual quality is closely related to the low-level image feature distortion  \cite{10.1167/9.10.1, Rad_2019,ping2008information,davis2015humans}, despite the fact that the explicit expression is intractable.} \revisionRG{Basically}{Generally speaking}, given an image, people tend to be more sensitive to the low-variation features instead of textures and edges \cite{devalois1990spatial,xia2012visual,karunasekera1995distortion,9195172}. Following the convention in image/video processing, people usually divide the low-level image features into three categories: smooth regions, edges, and textures. This \revisionRG{provides}{motivates the} \revisionRG{clue}{key idea of}  \revisionRG{quantifying}{quantitatively describing} the \revisionRG{perceptual}{feature} distortion of \revisionRG{adversarial}{the perturbed image in our work: assign different penalty to each feature categories based on their visual response sensitivity.}
    
	\revisionPQ{Here, the idea in this work is essentially in line with the core principle of the commonly-used JPEG image compression: we reduce or abandon the high-frequency components (small penalty) while keep the low-frequency parts (large penalty) after DCT transformation due to their distinct visual sensitivity. However, the strategy can not be directly used in producing adversarial images, due to the designed standards on image resolution and the inevitable blocking effects in JPEG compression \cite{wang2019cfsnet, farid2009exposing, jung2012image, 9195172}. Hence, we improve the initial strategy by assigning distinct penalties to different categories of low-level image feature distortion to maintain the \emph{spatial consistency and continuity} of the perturbed adversarial image. }{}
	
	\section{Perceptual Feature Fidelity Loss}
	As mentioned above, although the precise quantization of human visual evaluation is difficult to obtain, we can still find out the close connection between human visual perception and image feature distortion. To describe the feature distortion caused by adversarial perturbations, we introduce the FA (Fourier-Agrand) filter, which is the \emph{optimal steerable} filter, to distinguish smooth regions and fast-variation features based on the local directionality response (workable even under $-5\mbox{dB}$ PSNR). We further discriminate between the edges and textures empirically through the feature spatial sparsity \cite{wolfson1998examining,liu2018image}. (as shown in fig. \ref{fig:eg_vs_texture}). Finally, we propose a weighted feature loss based on the feature-sensitivity relationship with the obtained feature map, which leads to efficient and adaptive adversarial attack algorithms. 
	   		\begin{figure*}[h]
		\begin{center}
			\minipage{0.15\textwidth}
			\includegraphics[height=25mm, width=25mm]{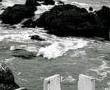}\par
			\centering reef
			\endminipage
			\minipage{0.15\textwidth}
			\includegraphics[height=25mm, width=25mm]{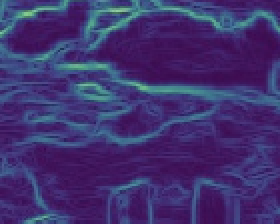}\par
			\centering FA response
			\endminipage
			\minipage{0.15\textwidth}
			\includegraphics[height=25mm, width=25mm]{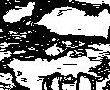}\par
			\centering smooth area
			\endminipage
			\minipage{0.15\textwidth}
			\includegraphics[height=25mm, width=25mm]{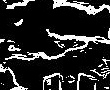}\par
			\centering edge
			\endminipage
			\minipage{0.15\textwidth}
			\includegraphics[height=25mm, width=25mm]{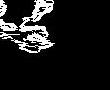}\par
			\centering texture
			\endminipage
		\end{center}
		\vspace{-1em}
		\caption{Example of FA classifier. For a given image (i.e., a reef image), we first compute the pixel-wise FA response, which gives rise to the FA response map (increase from cool to warm color). Then we first identify the smooth area whose FA response is small. Finally, we discriminate edges and textures based on spatial sparsity criterion. Note that the bright regions are features detected. }
		\label{fig:detected_features}
	\end{figure*}
	\subsection{Refined feature classification}\label{cls}
    To characterize the feature distortion of the perturbed image, the first and foremost step is to classify the low-level image features pixel by pixel robustly. This is a challenging job in image segmentation because of the diversity of real images. Over decades, people have proposed a series of detectors or classifiers to implement image segmentation, such as Canny edge detector \cite{canny1986computational}, deep-learning-based segmentation networks \cite{rad2019srobb}, steerable filter \cite{freeman1991design}, etc.. However, these methods cannot take both the computational efficiency and accuracy into account, resulting in rough and unsatisfactory results. Specifically, low-level features characterize the image local directionality, e.g., edges are usually unidirectional, yet textures are usually multidirectional. This requires that the \revisionRG{edge-detector}{classifier} used should be able to capture all possible directional changes, i.e., angles in $[0, 2\pi)$. However, it is very difficult to balance the accuracy and complexity: either we achieve finer angle discretization with expensive computational cost, or we keep low complexity by rougher angle discretization--as is done by the Canny edge detector. 
       	\begin{figure}[h]
    	\minipage{0.15\textwidth}
    	\includegraphics[width=\linewidth]{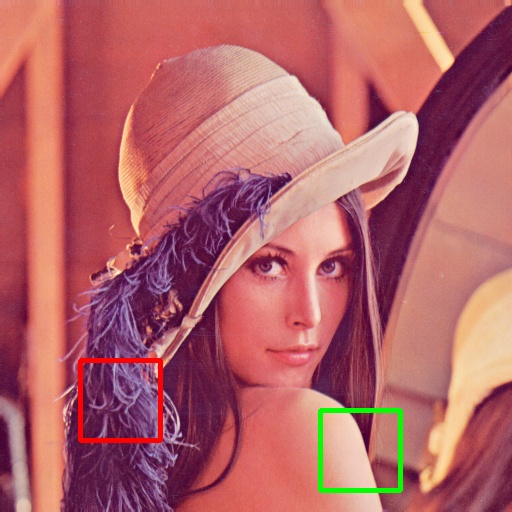} \centering (a) original\par
    	\endminipage\hfill
    	\minipage{0.15\textwidth}
    	\includegraphics[width=\linewidth]{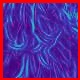} \centering (b) texture\par
    	\endminipage\hfill
    	\minipage{0.15\textwidth}
    	\includegraphics[width=\linewidth]{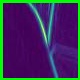} \centering  (c) edge\par
    	\endminipage\hfill
    	\caption{FA response of texture and edge. Fig. (a) corresponds to the red rectangle in the original image and is usually treated as texture. Fig. (b) corresponds to the green and is considered as edges. Clearly FA response of a texture region is denser than an edge region.}
    	\label{fig:eg_vs_texture}
    \end{figure}
			\begin{figure*}[h]
	\vspace{-1em}
	\centering
	\includegraphics[width=0.5\linewidth]{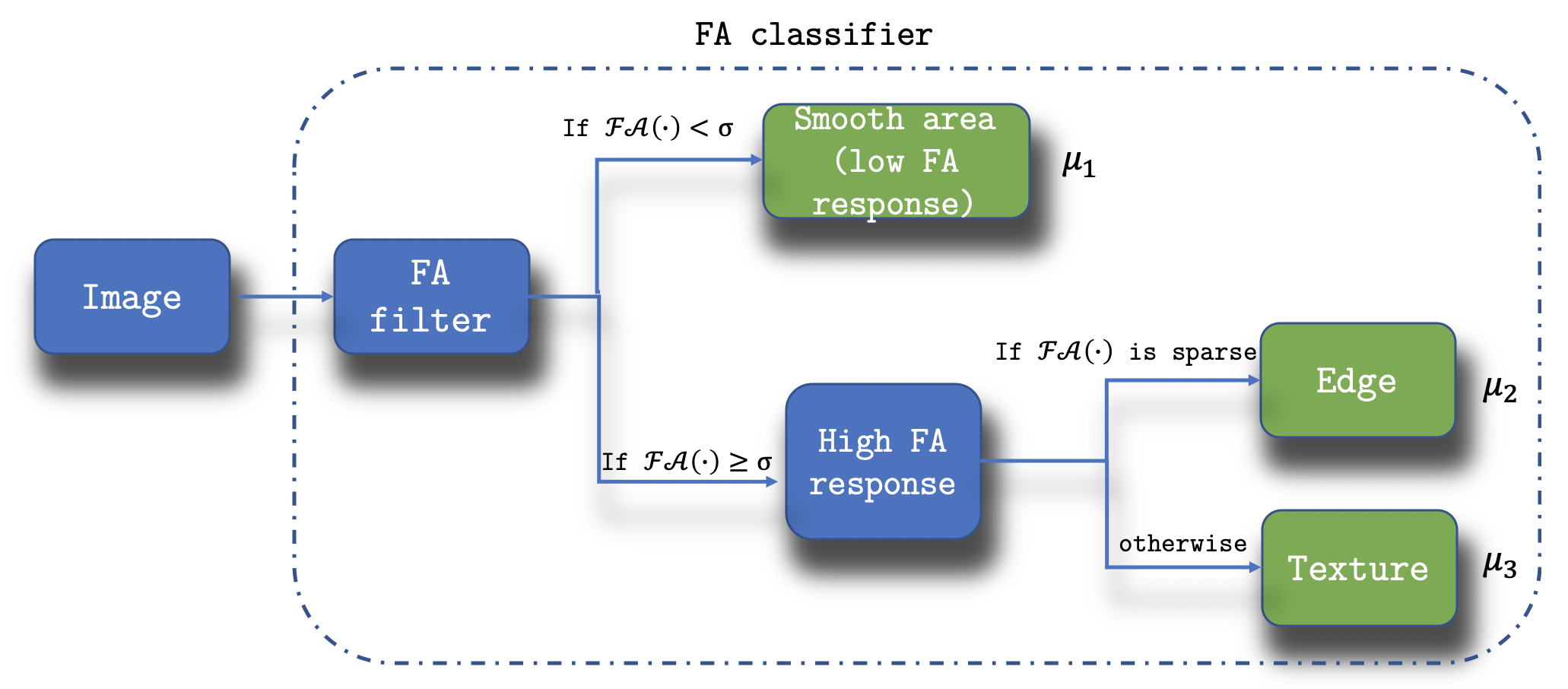}
	\caption{FA-filter based feature classification. For a pixel $x_{i,j}$, it will be first classified by the magnitude of FA directional response, i.e. $\mathcal{FA}(x_{i,j})$. If $\mathcal{FA}(x_{i,j}) < \sigma$, then the given pixel $x_{i,j}$ will be classified as the smooth area. Otherwise, the pixel will be further classified based on the spatial sparsity (Eqn. \ref{sparsity})}\label{procedure}
	\vspace{-1.5em}
\end{figure*}
	\par\noindent\textbf{FA Filter}. To handle the trade-off between computation and accuracy, people further propose steerable filters \cite{freeman1991design, zhao2020fourier}, whose space of all rotated versions is finite as long as the steerability assumption is satisfied. Using a linear combination of basis filters \cite{freeman1991design}, they can effectively obtain an approximated version of the original filter, which is rotation invariant. 
	\revisionPQ{}{Recently, people develop Fourier-Argand (FA) filter \cite{zhao2020fourier} and \revisionRG{guarantees}{prove} its \emph{optimality and  efficiency} in terms of the approximation error of steerable patterns.} \revisionPQ{}{To be more specific,} the key idea of the FA filter is to find the \textit{optimal} basis consisting of $N$ functions for approximating \textit{all} rotated versions of the given pattern (edge/texture) that is able to sensitively capture the image spatial variation along a certain direction. Specifically, let ${}^{\alpha} h$ denote the matching pattern characterized by the direction $\alpha$ ($\alpha \in [0, 2\pi]$), and $\{\phi_{0}, \phi_{1}, \cdots, \phi_{N - 1}\}$ be an arbitrary basis of $N$ elements. Let $\mathcal{P}\{\cdot \}$ denote the orthogonal projection onto the approximation space $span\{\phi_{0}, \phi_{1}, \cdots, \phi_{N - 1}\} $. Zhao et al. \cite{zhao2020fourier} find the \emph{optimal basis} for all rotated versions of ${}^{\alpha} h$ by minimizing the average approximation error \revisionRG{arg1}{$e_{N}^{2}$ over all possible angles}:
	\begin{align}
	\revisionRG{arg1}{e_{N}^{2}} \overset{\mathrm{def}}{=} \int_{0}^{2\pi} \norm{{}^{\alpha} h- \mathcal{P}\{{}^{\alpha} h\}}^{2}\; d\alpha
	\end{align}
	This minimization automatically leads to the optimal and rotation-invariant Fourier-Argand basis $\phi_{i}^{*}$. The optimality and the rotation invariance ensure that all spatial directions can be accurately characterized with minimal angle quantization error. The properties provide an efficient tool to accurately identify the fast-variation and directional image features with small computational cost, and hence is the key of the proposed image distortion metric. We refer to this paper \cite{zhao2020fourier} for more details if readers have interests.

	\par\noindent\textbf{Low-level Feature Classification.} The next question is to further discriminate the edge and texture features, given the FA directionality response. The key idea is based on the following observation: intuitively, the spatial sparsity of the texture features in the FA response is much higher than the edge features as shown in fig. \ref{fig:eg_vs_texture}. \revisionPQ{}{This provides a valid criterion to effectively \revisionRG{classify}{discriminate} the edge and texture features which has been used for decades \cite{liu2018image, wolfson1998examining, texture1, texture2}. We need to point out that this texture discrimination is tentative and empirical from image processing tasks. So far, similar to human visual evaluation, there is still no clear definition to discriminate edge and texture \cite{humeau2019texture}, and we are actively working towards better criterion. Fig. \ref{fig:detected_features} gives an intuitive example of low-level image feature classification. }

	To be more specific, let $\mu_1$, $\mu_2$, $\mu_3$ denote three \revisionRG{human perception coefficients}{distortion penalty of} smooth \revisionRG{area}{region}, edge, and texture respectively. Let $\mathbf{1_{\{\cdot\}}}$ \revisionRG{is}{denote the} indicator function. We first define \revisionRG{a sparsity function $g(x_{i,j})$ to characterize the}{the} sparsity of FA \revisionRG{response}{directional response} within a local patch $B_{i,j}$ centered at pixel $x_{i,j}$:

	\begin{equation}
	B_{i, j}:  \overset{\mathrm{def}}{=} \{x_{m,n}\mid \revisionRG{i-r_0\leq m \leq i+r_0}{|m - i| < r_{0}}, \; \revisionRG{j-r_0\leq n \leq j+r_0}{|n - j| < r_{0}}\}
	\end{equation}
	
	\begin{equation}\label{sparsity}
	g(x_{i,j}) = \frac{1}{|B_{i, j}|}\sum_{x_{m,n}\in B_{i, j}} \mathbf{1}_{\{\mathcal{FA}(x_{m,n}) \geqslant \sigma\}}
	\end{equation}
	therefore, $g(x_{i,j})$ \revisionRG{will compute}{reflects the directional response density} within the \revisionRG{patch}{local patch} $B_{i,j}$. With the response density $g(x_{i,j})$, the low-level image features can be tentatively discriminated: for pixel with FA response smaller than the high-variation threshold $\sigma$, it will be classified as the smooth area category; for the others, the pixel will either be classified as texture if its FA directional response density is larger than $s_{0}$, or edge category if its response density is smaller than $s_{0}$. Mathematically, the feature category label is given by:
	\begin{equation}\label{M}
	\begin{aligned}
	& &M_{i,j} = \left\{\begin{array}{ll}
	\mu_1, & \mathcal{FA}(x_{i,j}) < \sigma \\
	\mu_2 , & \mathcal{FA}(x_{i,j}) \geqslant \sigma \text{ and }  g(x_{i,j}) \leqslant s_{0}\\
	\mu_3 , &  otherwise\\
	\end{array}
	\right.\\
	\\
	\end{aligned}
	\end{equation}
	where $M_{i,j}$ indicates the sensitivity \revisionRG{coefficient}{penalty} of pixel $x_{i,j}$ \revisionRG{arg1}{depending on its feature category. Fig.~\ref{procedure} gives an intuitive depiction of the flowchart of the FA classifier.} 
	
	\subsection{Optimization Formulation \& Algorithmic Framework }
	\noindent\textbf{Perceptual Feature Fidelity Loss.} Consider the input image $\mathbf{x} \in \mathbb{R}^N$ where $N$ is the size of the image. The hard-label classifier gives its predicted label $y$, where $y\in\{1,...,C\}$. Given the original image $\mathbf{x}^{org}$, its ground-truth label $y^{org}$, a target class $t \neq y^{org}$, we generate the adversarial image $\mathbf{x}^{adv}$ using Perceptual Feature Fidelity Loss (PFFL):
	\begin{subequations}\label{L2_loss}
		\begin{alignat}{2}
		&\underset{\mathbf{x}^{adv}\in \mathbb{R}^{N}}{\text{minimize}}& \qquad & PFFL(\mathbf{x}^{org}, \mathbf{x}^{adv})\\
		&\text{subject to} & &f(\mathbf{x}^{adv})=t
		\end{alignat}
	\end{subequations}
	where
	\begin{equation}\label{perc_loss}
	PFFL(\mathbf{x}^{org}, \mathbf{x}^{adv}): \overset{\mathrm{def}}{=} ||\mathbf{x}^{org}\odot \boldsymbol{M} -  \mathbf{x}^{adv} \odot \boldsymbol{M}||_2^2
	\end{equation}

The notation $\odot$ denotes the Hadamard product operator. PFFL strives to quantitatively describe the low-level image feature distortion with the weighted sensitivity penalty $\boldsymbol{M}$.

\noindent\textbf{Proposed Algorithm.} Algorithm \ref{algo_sign_opt} gives the implementation details to illustrate the attack process. Note that the function $g(\boldsymbol{\theta})$ is proposed in \cite{cheng2019sign, cheng2018query}, where the hard-label black-box attack is formulated as a real-valued optimization problem $g(\boldsymbol{\theta})$. Except for the SignOPT realization, we also leverage HSJA \cite{chen2020hopskipjumpattack} and Boundary Attack \cite{brendel2017decision} to demonstrate the efficiency and flexibility of PFFL (for more detailed implementation and experiments, we refer our reader to the appendix).

\begin{algorithm}[h]
    \caption{PFFL + \text{SignOPT}}
    \label{algo_sign_opt}
    \begin{algorithmic}[1]
    \State Given original image $\mathbf{x}^{org}$, image $\mathbf{x}^{target}$ in the target class $t$, hard-label black-box classifier $f(\mathbf{x}):\mathbb{R}^N \rightarrow \{0,1,...,C\}$
    \State Define function $g(\boldsymbol{\theta}):\overset{\mathrm{def}}{=}\mathop{argmin }_{\lambda > 0} {\text{ s.t. } f(\mathbf{x}^{org}+\lambda\frac{\boldsymbol{\theta}}{||\boldsymbol{\theta} \odot \boldsymbol{M}||}) = t}$
    \State Generate $\boldsymbol{M}\in \mathbb{R}^N$ according to the procedures described in the paper. Let $\boldsymbol{\theta}_0 = \mathbf{x}^{target} - \mathbf{x}^{org}$
    \For{\texttt{$i=0:N_0$}}
        \State Generate random noise $\boldsymbol{\eta}_1, \boldsymbol{\eta}_2, \cdots, \boldsymbol{\eta}_Q  \in \mathbb{R}^N$ from Gaussian or Uniform distribution
        \For{\texttt{$q=1:Q$}}
            \State $\boldsymbol{\eta}_q \leftarrow (\boldsymbol{\eta}_q \odot \frac{1}{\boldsymbol{M}})^2$
        \EndFor
        \State Compute $\nabla \hat{g}(\boldsymbol{\theta}_i) = \frac{1}{Q} \sum_{q=1}^{Q}{\text{sign}(g(\boldsymbol{\theta}_i + \epsilon \boldsymbol{\eta}_q)-g(\boldsymbol{\theta}_i))\cdot\boldsymbol{\eta}}_q$
        \State Choose an appropriate step size $\gamma$ using line search
        \State Update $\boldsymbol{\theta}_{i+1} = \boldsymbol{\theta}_{i} - \gamma \nabla \hat{g}(\boldsymbol{\theta}_i)$
    \EndFor
    \State \textbf{return } $\mathbf{x}^{org} + g(\boldsymbol \theta_{i+1})\boldsymbol{\theta}_{i+1}$
    \end{algorithmic}
\end{algorithm}

	\begin{table*}[!htbp]
		\centering
		\begin{adjustbox}{max width=0.7\textwidth}
			\begin{tabular}{l|lll|lll}
				\toprule
				SSIM /PSNR   & \multicolumn{3}{c}{SignOPT}     & \multicolumn{3}{c}{HSJA} \\
				\midrule
				Queries & PFFL & $L_2$ & $L_\infty$ & PFFL & $L_2$ & $L_\infty$ \\
				\midrule
				4.8k & \textbf{0.60} /\textbf{23.15} & 0.46 /22.36 & 0.38 /18.63 & \textbf{0.50} /19.95 & 0.30 /\textbf{20.11} & 0.18 /17.44 \\
				\midrule
				7.2k & \textbf{0.71} /\textbf{25.75} & 0.54 /25.03 & 0.41 /19.38 & \textbf{0.63} /21.70 & 0.38 /\textbf{22.60} & 0.23 /19.27 \\
				\midrule
				9.6k & \textbf{0.77} /\textbf{27.67} & 0.61 /27.24 & 0.44 /19.89 & \textbf{0.74} /23.73 & 0.47 /\textbf{25.08} & 0.29 /20.37 \\
				\midrule
				12k & \textbf{0.83} /\textbf{29.55} & 0.66 /29.31 & 0.47 /20.47 & \textbf{0.80} /25.07 & 0.55 /\textbf{26.99} & 0.34 /21.34 \\
				\bottomrule
			\end{tabular}
		\end{adjustbox}
		\caption{Performance comparison on targeted attacks on ResNet-50. Column: objective function (such as PFFL, $L_2$, and $L_\infty$) + attack algorithm. Row: queries. Obviously, our method achieve the best SSIM with $0.2 +$ improvements in all cases and the best (in SignOPT) or comparable (in HSJA) PSNR performance. Usually, $\mathbf{0.1 +}$ SSIM increment under similar PSNR values means significant visual quality improvement in image processing \cite{wang2019cfsnet, wang2004image}.  }
		\label{table:diff attacks}
	\end{table*}

\begin{table*}[h]
	\centering
	\begin{adjustbox}{max width=0.7\textwidth}
		\begin{tabular}{l|lll|lll}
			\toprule
			SSIM /PSNR   & \multicolumn{3}{c}{DenseNet} & \multicolumn{3}{c}{MobileNet} \\
			\midrule
			Queries & PFFL & $L_2$ & $L_\infty$ & PFFL & $L_2$ & $L_\infty$ \\
			\midrule
			4.8k & \textbf{0.60} /\textbf{22.83} & 0.44 /22.62 & 0.35 /18.24 & \textbf{0.60} /22.42 & 0.49 /\textbf{22.95} & 0.39 /19.25\\
			\midrule
			7.2k & \textbf{0.70} /25.66 & 0.53 /\textbf{25.93} & 0.40 /18.73 & \textbf{0.69} /25.10 & 0.60 /\textbf{26.06} & 0.43 /20.63 \\
			\midrule
			9.6k & \textbf{0.77} /27.79 & 0.61 /\textbf{28.55} & 0.43 /19.53 & \textbf{0.77} /27.44 & 0.67 /\textbf{28.62} & 0.45 /21.08 \\
			\midrule
			12k & \textbf{0.82} /29.50 & 0.68 /\textbf{30.00} & 0.45 /20.06 & \textbf{0.81} /29.30 & 0.71 /\textbf{30.64} & 0.47 /21.81 \\
			\bottomrule
		\end{tabular}
	\end{adjustbox}
	\caption{Performance comparison on different networks using SignOPT. Column: objective function (such as PFFL, $L_2$, and $L_\infty$). Row: queries. Our method achieve the best SSIM with at $0.1 + $  increments and comparable PSNR performance. It effectively demonstrates the flexibility of PFFL that can be easily incorporated into various network architectures.}
	\label{table:diff nets}
	\vspace{-1em}
\end{table*}

The main difference from \cite{cheng2019sign} is that the objective function is replaced with $PFFL(\mathbf{x}^{org}, \mathbf{x}^{adv})=\|(\mathbf{x}^{org}-\mathbf{x}^i)\odot \boldsymbol{M}\|_2$. Furthermore, we also conduct untargeted hard-label black-box attack, where the starting image is initialized with:
\begin{equation}
    \mathbf{x}^{target} = (\boldsymbol{\eta}\odot\frac{1}{\boldsymbol M})^2 \text{ s.t. } f(\mathbf{x}^{target})\neq y^{org}
\end{equation}
where random noise $\boldsymbol{\eta} \in \mathbb{R}^N$ is generated from Gaussian or uniform distribution. The other procedures remain the same.

   	\begin{figure*}[h]
   	\begin{center}
		\minipage{0.14\textwidth}
			\includegraphics[width=0.95\textwidth]{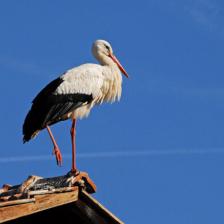}\par
		\endminipage
		\minipage{0.14\textwidth}
			\includegraphics[width=0.95\textwidth]{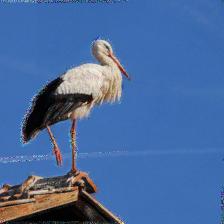}\par
		\endminipage
		\minipage{0.14\textwidth}
			\includegraphics[width=0.95\textwidth]{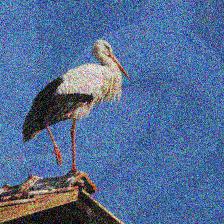}\par
		\endminipage
		\minipage{0.14\textwidth}
			\includegraphics[width=0.95\textwidth]{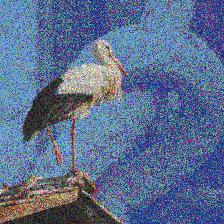}\par
		\endminipage
        \minipage{0.14\textwidth}
			\includegraphics[width=0.95\textwidth]{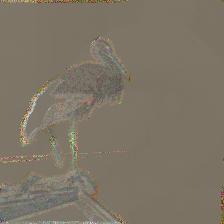}\par
		\endminipage
        \minipage{0.14\textwidth}
        \includegraphics[width=0.95\textwidth]{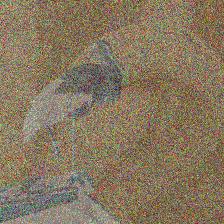}\par
		\endminipage
        \minipage{0.14\textwidth}
			\includegraphics[width=0.95\textwidth]{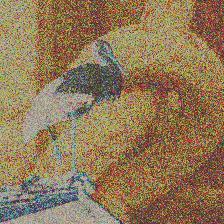}\par
		\endminipage
		
		\minipage{0.14\textwidth}
			\includegraphics[width=0.95\linewidth]{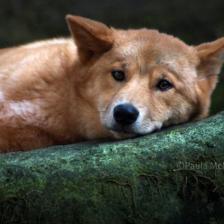}\par
			\centering original image
		\endminipage
		\minipage{0.14\textwidth}
			\includegraphics[width=.95\linewidth]{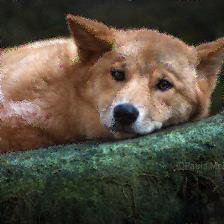}\par
			\centering PFFL
		\endminipage
		\minipage{0.14\textwidth}
			\includegraphics[width=.95\linewidth]{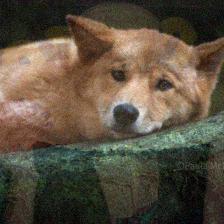}\par
			\centering $L_2$
		\endminipage
		\minipage{0.14\textwidth}
			\includegraphics[width=.95\linewidth]{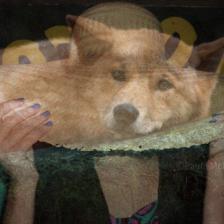}\par
			\centering $L_\infty$
		\endminipage
		\minipage{0.14\textwidth}
			\includegraphics[width=0.95\textwidth]{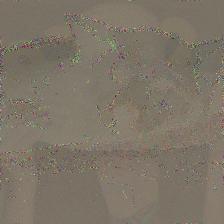}\par
			\centering PFFL (noise)
		\endminipage
        \minipage{0.14\textwidth}
			\includegraphics[width=0.95\textwidth]{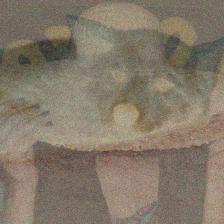}\par
			\centering $L_2$ (noise)
		\endminipage
        \minipage{0.14\textwidth}
			\includegraphics[width=0.95\textwidth]{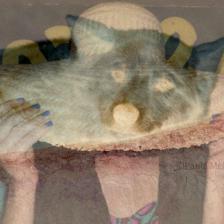}\par
			\centering $L_\infty$ (noise)
		\endminipage
	\end{center}
	\vspace{-1em}
	\caption{Visualization examples of targeted attacks with $5$k iterations and the corresponding noise injected. 
	Notice that the significant ghost effects and salt and pepper noise generated by $L_{2}$ and $L_{\infty}$ based methods. Besides, more noise are encouraged in edge or texture region to improve the imperceptibility using PFFL. Wth limited queries, optimizing PFFL are more effective to generate imperceptible perturbation than other metrics.}\label{fig:targeted}
\end{figure*}

\section{Experimental results}
\begin{figure*}[!tbh]
	\minipage{0.16\textwidth}
	\includegraphics[width=\textwidth]{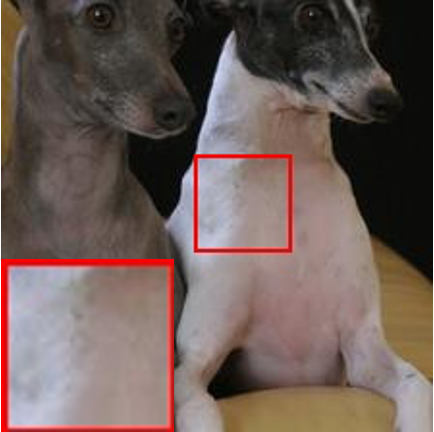}\par
	\centering \small original
	\endminipage\hfill
	\minipage{0.16\textwidth}
	\includegraphics[width=\textwidth]{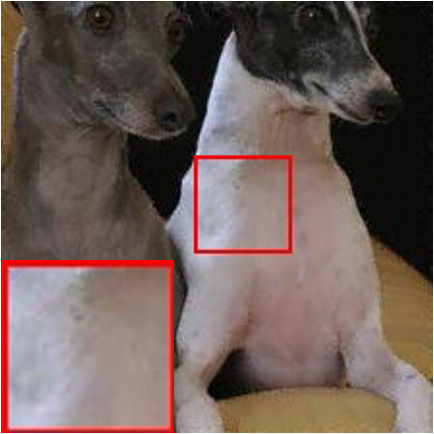}\par
	\centering \small PFFL+HSJA
	\endminipage\hfill
	\minipage{0.16\textwidth}
	\includegraphics[width=\textwidth]{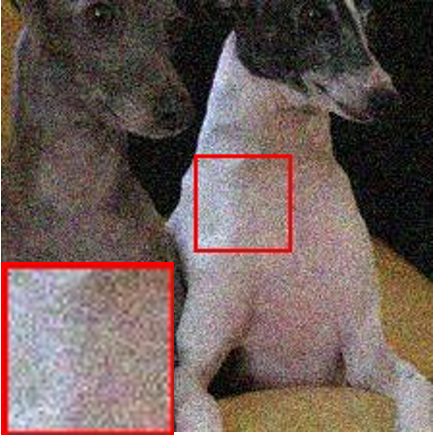}\par
	\centering \small $L_2$+HSJA
	\endminipage\hfill
	\minipage{0.16\textwidth}
	\includegraphics[width=\textwidth]{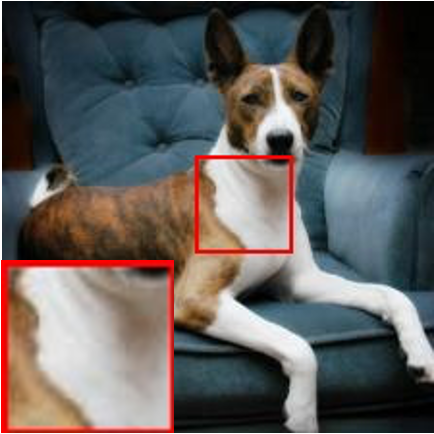}\par
	\centering \small original
	\endminipage\hfill
	\minipage{0.16\textwidth}
	\includegraphics[width=\textwidth]{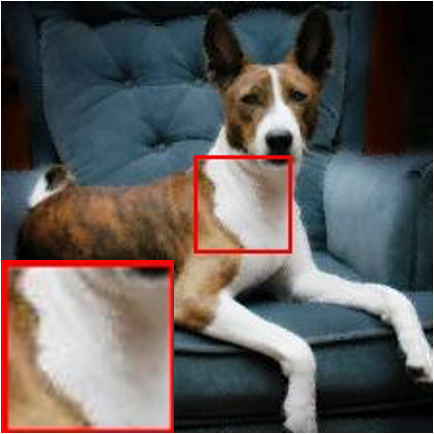}\par
	\centering \small PFFL+SignOPT
	\endminipage\hfill
	\minipage{0.16\textwidth}
	\includegraphics[width=\textwidth]{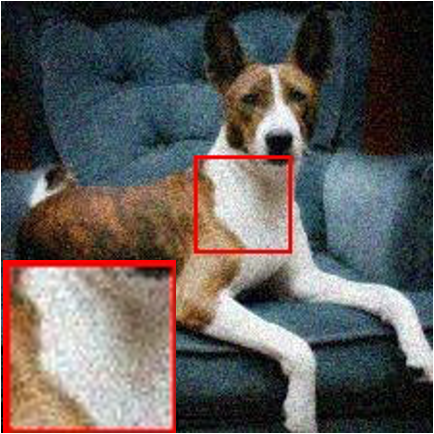}\par
	\centering \small $L_2$+SignOPT
	\endminipage\hfill
	\vspace{-0.5em}
	\caption{Visualization examples of untargeted attack with $1$k queries. 
	Notice that optimizing $L_{2}$ gives rise to significant undesired grain effects on the whole image, while PFFL leads to satisfactory results with smoother objects and sharper details. This intuitively shows that optimizing PFFL is much more efficient and robust in query-limited adversarial attacks.  }\label{fig:untargeted}
	\vspace{-0.5em}
\end{figure*}

	\noindent\textbf{Experiment setup.} We first randomly select 100 correctly classified image pairs from ImageNet test dataset\cite{deng2009imagenet}. Then we use SignOPT \cite{cheng2019sign} and HopSkipJumpAttack (HSJA)\cite{chen2020hopskipjumpattack} frameworks to optimize the PFFL objective. Experiments are conducted on three different network architectures: ResNet-50\cite{he2016deep}, DenseNet\cite{huang2017densely}, and MobileNet\_v2\cite{sandler2018mobilenetv2}. We mainly focus on the hard-label black-box attack setting, where the initial images are correctly classified in the targeted class. Besides, $\mu_1$, $\mu_2$, and $\mu_3$ in Eqn. \eqref{M} are set to 1, 0.3, and 0.5 respectively. $r_0$ is 1/10 of the width of input images and $s_0$ equals to 0.4. In the experiments, each input image is normalized following the conventions used in PyTorch\footnote{\url{https://pytorch.org/docs/stable/torchvision/models.html}}. PFFL is calculated as described in \eqref{perc_loss}. To evaluate the quality of generated images, we mainly use SSIM\cite{wang2004image} and PSNR\eqref{eq:PSNR} as our referring metrics. We report the median value of PSNR and SSIM across all examples within a specific query in the following experiments.
	\begin{equation}\label{eq:MSE}
	    MSE = \frac{1}{N}\sum_{i=1}^{N}(x^{org}_i-x^{adv}_i)^2
	\end{equation}
	
	\begin{equation}\label{eq:PSNR}
	    PSNR = 10log_{10}(\frac{MAX_I^2}{MSE})
	\end{equation}
	where $MAX_I$ is the maximum possible pixel value of the image.

\subsection{Comparison on different attacks}

In these experiments, we conduct attacks using different objectives (PFFL, $L_2$, and $L_\infty$) and attack frameworks (SignOPT and HSJA). We calculate their corresponding SSIM and PSNR value as indicators of visual quality at different queries. From table \ref{table:diff attacks}, we can observe a significant improvement in SSIM (generally 0.1+ SSIM increase already means significant visual quality improvement \cite{wang2019cfsnet, wang2004image}) and a comparable PSNR of PFFL perturbed adversarial images using the same amount of queries. Moreover, the adversarial attacks using PFFL produce images with much higher SSIM and PSNR than the attack using $L_\infty$. This quantitatively demonstrates that PFFL can effectively guide the noise distribution to improve visual quality without changing the overall noise power. Furthermore, fig. \ref{fig:targeted} visually shows the comparison using different metrics with the same amount of query and demonstrates that adversarial perturbations generated using PFFL are indeed more imperceptible given the benign noise distribution. Here we need to point out that when the noise power is too small, our improvement is not as drastic because the injected noise is nearly unnoticeable. However, when the noise power is non-trivial, the visual gain produced by benign noise distributions is apparent. For more visualization examples, we refer our readers to Section 1 of the appendix. Additionally, a similar result can be observed in the experiments implemented on different network architectures, as shown in table \ref{table:diff nets}. The above results indicate that PFFL can be flexibly applied to other network structures and attack frameworks. 

For the untargeted attacks, we only report results generated using fewer query budget, as the perturbation is already imperceptible in the later iteration indicated by the high PSNR and SSIM. In this setting, the improvement of PFFL is also remarkable, as shown by table \ref{table:diff untargeted attack}. When the query budget is further limited, e.g., using 800 queries, PFFL generates adversarial perturbations, which has much better visual quality with considerable $0.35 +$ SSIM and $2 + \mbox{dB} $ PSNR improvements. Figure \ref{fig:untargeted} intuitively demonstrates the significant visual quality improvement by using PFFL to strategically guide the perturbation spatial distribution, in contrast to the undesired salt and pepper noise in $L_p$-based examples.
	\begin{figure*}[!tbh]
	\minipage{0.14\textwidth}
	\includegraphics[width=\linewidth]{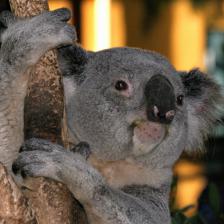}\par
	\centering original
	\endminipage\hfill
	\minipage{0.14\textwidth}
	\includegraphics[width=\linewidth]{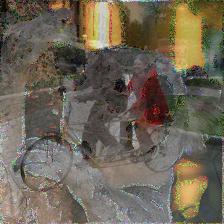}\par
	\centering PSNR=17.23
	\endminipage\hfill
	\minipage{0.14\textwidth}
	\includegraphics[width=\linewidth]{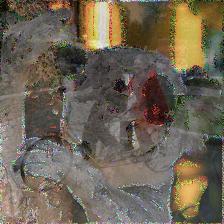}\par
	\centering PSNR=18.01
	\endminipage\hfill
	\minipage{0.14\textwidth}
	\includegraphics[width=\linewidth]{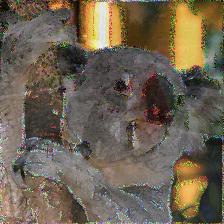}\par
	\centering PSNR=19.47
	\endminipage\hfill
	\minipage{0.14\textwidth}
	\includegraphics[width=\linewidth]{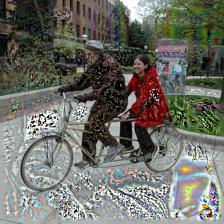}\par
	\centering PSNR=10.17
	\endminipage\hfill
	\minipage{0.14\textwidth}
	\includegraphics[width=\linewidth]{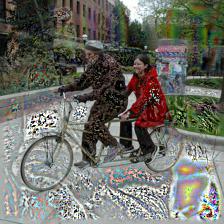}\par
	\centering PSNR=10.21
	\endminipage\hfill
	\minipage{0.14\textwidth}
	\includegraphics[width=\linewidth]{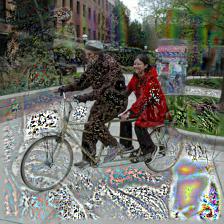}\par
	\centering PSNR=10.23
	\endminipage\hfill
	\caption{PSNR of PFFL and SSIM when used as the attack objectives. The second to forth images are produced using PFFL with 0.5k, 1k, and 2k queries. The last three images are generated using SSIM with the same query budget as PFFL. Suppressing the overall distortion is an important factor for visual quality while SSIM emphasizes too much on restoring the object local structure.}
	\label{fig:comp_PFFL_SSIM}
\end{figure*}	

Besides, we also conduct human perception evaluation on adversarial images generated by different metrics (results are reported in the appendix). The human perception study also supports that our proposed approach can generate much more imperceptible adversarial examples than baseline methods.
	\begin{table}[h]
        \centering
        \begin{adjustbox}{max width=0.49\textwidth}
            \begin{tabular}{l|ll|ll}
                \toprule
                SSIM/PSNR   & \multicolumn{2}{c}{SignOPT}     & \multicolumn{2}{c}{HSJA} \\
                \midrule
                Queries & PFFL & $L_2$ & PFFL & $L_2$ \\
                \midrule
                0.8k & \textbf{0.67} /\textbf{25.42}& 0.32 /23.00 & \textbf{0.65} /\textbf{20.41} & 0.20 /18.64 \\
                \midrule
                1.6k & \textbf{0.77} /\textbf{28.09}& 0.45 /26.40& \textbf{0.87} /28.28 & 0.57 /\textbf{29.72} \\
                \midrule
                2.4k & \textbf{0.88} /\textbf{32.12}& 0.63 /30.99& \textbf{0.92} /31.65 & 0.78 /\textbf{34.27} \\
                \midrule
                3.2k & \textbf{0.94} /\textbf{35.64}& 0.74 /34.48& \textbf{0.94} /33.23 & 0.85 /\textbf{37.30} \\
                \bottomrule
            \end{tabular}
        \end{adjustbox}
        \caption{Performance comparison on untargeted attacks. Our method can achieve superior PSNR and SSIM performance that are much higher than the $L_{2}$-based technique, where the PSNR and SSIM increments are $2+ \mbox{dB}$ and 0.35 +, especially using fewer queries ($< 1000$). This makes PFFL particularly useful in practical applications with query restrictions. }\label{table:diff untargeted attack}
    \end{table}

\subsection{Why PFFL is a suitable objective?}
In order to get a deeper insight into the effectiveness of PFFL, we further compare the adversarial images generated using SSIM as the attack objective. We optimize the SSIM metric using the HSJA algorithm. From fig. \ref{fig:comp_PFFL_SSIM}, we \revisionRG{observe}{can find} that images produced have much worse image quality with visible undesired textures and patterns. Notice that the PSNR performance of the SSIM-based adversarial examples is quite poor because the SSIM metric does not consider the overall noise power, as discussed in Sec. \ref{background}. This again shows that optimizing the existing metric (SSIM, $L_2$ $L_{\infty}$, etc.) solely can hardly give rise to an efficient and visually satisfactory result. 

The superior imperceptibility of images produced by PFFL \revisionRG{may be}{is} endowed by \revisionRG{arg1}{suppressing the overall noise power and local feature distortion simultaneously.} To get more insights into the characteristics of PFFL, we further conduct a study on the correlation between PFFL, PSNR, and SSIM without attacks. We start from images with noise of different power added to it. By using projected gradient descent, we decrease the PFFL while making either PSNR or SSIM unchanged. From table \ref{table:correlation} we can observe \revisionRG{a strong correlation between PFFL, SSIM, and PSNR}{that}: PFFL monotonically increases \revisionRG{when}{as} SSIM decreases and PSNR remains unchanged; PFFL \revisionRG{also behaves similarly}{monotonically increases} when PSNR reduces and SSIM is fixed. \revisionRG{While SSIM focuses on the local structure similarity of local regions and PSNR evaluates the distortion as a whole without pixel-wise differentiation}{This effectively shows that}, PFFL consider both aspects simultaneously when evaluating image quality and thus is a more reliable \revisionRG{approach}{measure}.

\begin{figure}[!htb]
    \centering \minipage{0.236\textwidth}
        \includegraphics[width=\linewidth]{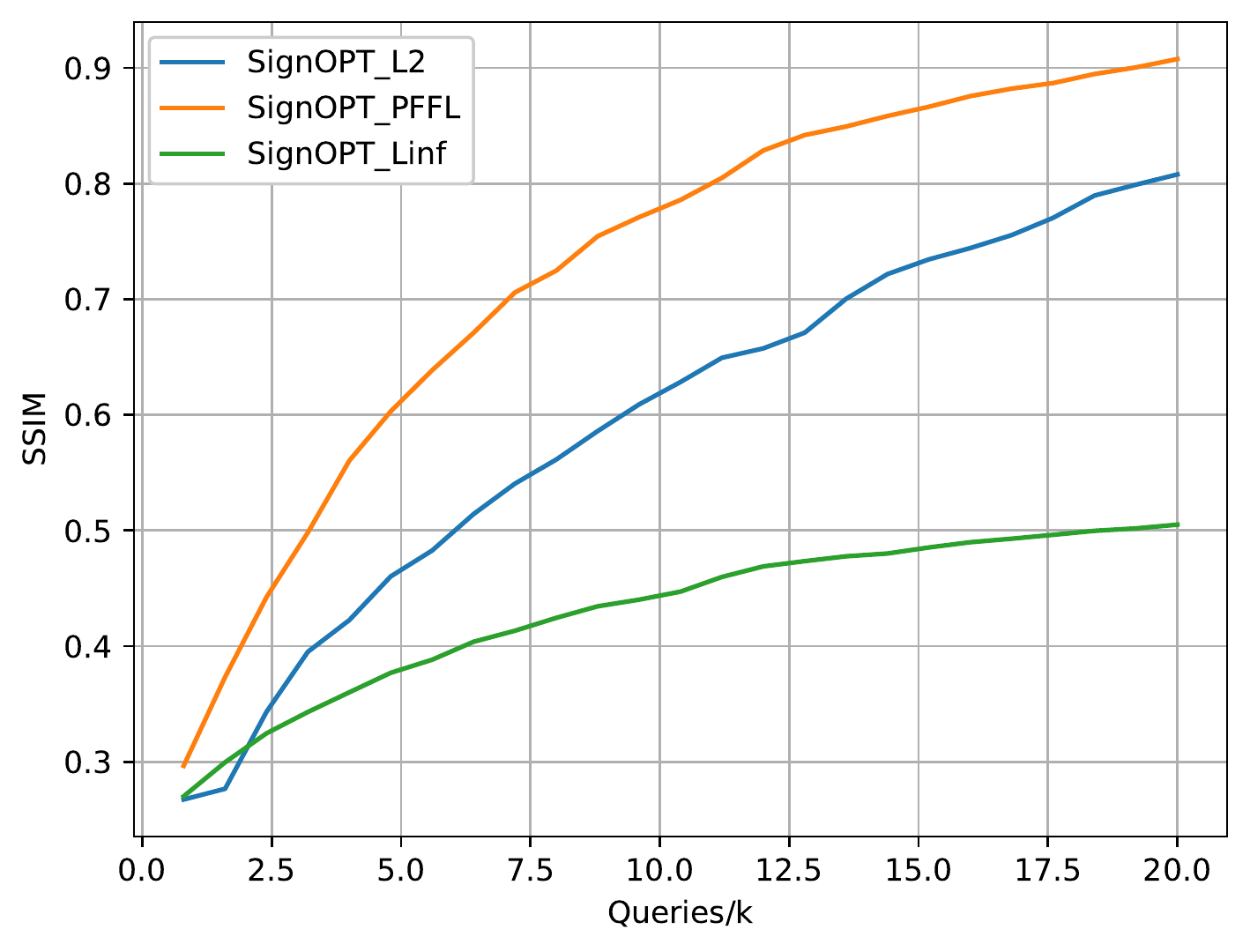}\par
    \endminipage\hfill
    \centering \minipage{0.236\textwidth}
        \includegraphics[width=\linewidth]{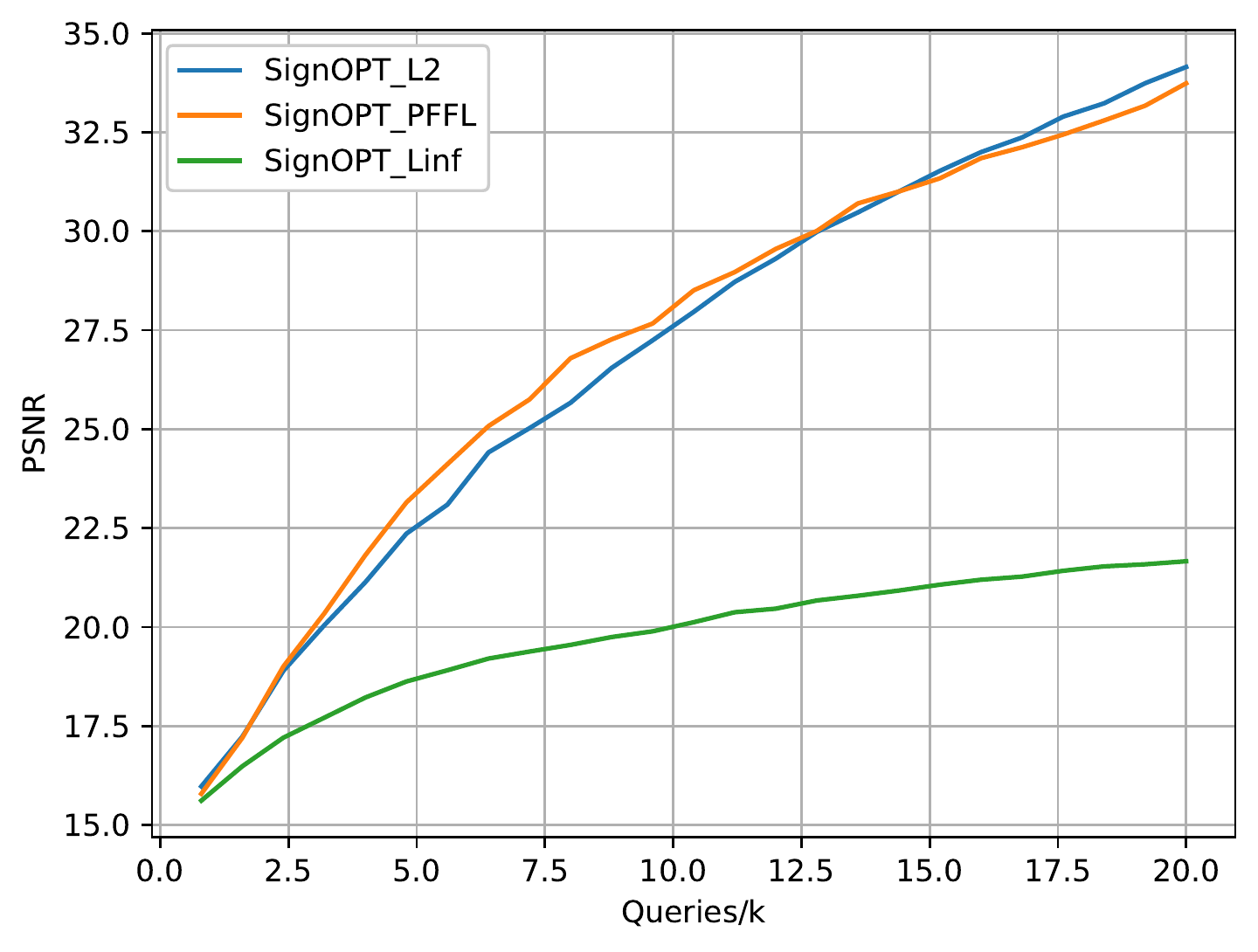}\par
    \endminipage\hfill
		\caption{Performance v.s. queries using SignOPT. With the same perturbation power (comparable PSNR in the right plot), PFFL leads to a much better visual quality with average $0.17$ SSIM increment over $L_{2}$-based method. This effectively demonstrates the superiority of the query-efficiency of PFFL.  }
	\label{fig:performance}
	\end{figure}

	\begin{table}[h]
		\centering
		\begin{adjustbox}{max width=0.48\textwidth}
            \begin{tabular}{l|l| l l l l l l l}
				\toprule
				 \multicolumn{2}{c}{\diagbox{PFFL}{PSNR}} & 19.0 & 17.0 & 14.5 & 13.0 & 10.0 & 8.22 & 6.46 \\
				\midrule
				\multirow{6}{*}{SSIM} & 0.7 & 66.13 & 76.07 & 97.65 & 116.3 & - & - & - \\ 
                & 0.6 & 82.33 & 90.35 & 107.3 & 122.5 & 166.5 & 202.5 & 247.3 \\ 
                & 0.5 & 105.9 & 106.8 & 121.1 & 136.5 & 180.6 & 215.5 & 258.0 \\ 
                & 0.4 & 143.1 & 130.5 & 140.0 & 152.2 & 193.9 & 229.4 & 274.1 \\ 
                & 0.3 & 175.1 & 171.3 & 166.3 & 178.4 & 213.5 & 247.6 & 290.9 \\ 
                & 0.2 & - & 191.5 & 213.6 & 218.2 & 249.6 & 281.8 & 322.5 \\
				\bottomrule
			\end{tabular}
		\end{adjustbox}
		\caption{The relationship between PFFL, PSNR, and SSIM without attack. Column: varying PSNR values. Row: SSIM. The blank cells indicate that the corresponding value cannot be attained using projected gradient descent. Under the same PSNR value, PFFL decreases monotonically w.r.t. SSIM. Similarly, PFFL also reduces as PSNR values rises, given a fixed SSIM. It shows that PFFL is a comprehensive metric covering both PSNR and SSIM. }
	\label{table:correlation}
	\end{table}
	From fig. \ref{fig:performance} where different objective functions are used as the attack objective, it is straightforward to see that PFFL give rise to the perturbed adversarial results with higher SSIM and comparable PSNR, even though the $L_{2}$ metric directly optimize PSNR index. Better image quality is \revisionRG{attained}{efficiently attained} using PFFL through correlating perturbations with local image features and \revisionRG{suppressing the total distortion}{overall noise power}. This is in line with our \revisionRG{discussion}{motivation as discussed in Sec. \ref{background}.}
	\section{Conclusion}
	In this work, we propose a new metric, Perceptual Feature Fidelity Loss, for adversarial attack based on low-level image features for better visual quality. The metric relies on our novel low-level feature classifier and is compatible with different optimization methods. With the total distortion amount ($L_2$) comparable, our method can strategically change the noise distribution to improve imperceptibility, which is also verified quantitatively and qualitatively by our experiments.

\section*{Acknowledgement}
The research presented in this paper is supported in part by  the U.S. Army Research Laboratory and the U.K. Ministry of Defence under Agreement Number W911NF-16-3-0001 and the National Science Foundation (NSF) under awards \# CNS-1705135 and CNS-1822935, and by the National Institutes of Health (NIH) award \# P41EB028242 for the mDOT Center. The views and conclusions contained in this document are those of the authors and should not be interpreted as representing the official policies, either expressed or implied, of the NSF, the NIH, the U.S. Army Research Laboratory, the U.S. Government, the U.K. Ministry of Defence or the U.K. Government. The U.S. and U.K. Governments are authorized to reproduce and distribute reprints for Government purposes notwithstanding any copyright notation hereon.

{\small
\bibliographystyle{ieee_fullname}
\bibliography{egbib}
}

\end{document}